\documentclass{article}
\usepackage{ijcai25}

\usepackage{times}
\usepackage{soul}
\usepackage{url}
\usepackage[hidelinks]{hyperref}
\usepackage[utf8]{inputenc}
\usepackage[small]{caption}
\usepackage{graphicx}
\usepackage{amsmath}
\usepackage{amssymb}
\usepackage{amsthm}
\usepackage{enumitem}
\usepackage{booktabs}
\usepackage{algorithm}
\usepackage{multirow}
\usepackage[switch]{lineno}
\usepackage{algpseudocode, xcolor}

\newcommand{\methodname}{{\tt{FedDDL}}}

\title{Federated Deconfounding and Debiasing Learning for Out-of-Distribution Generalization}
\author{
Zhuang Qi$^1$, 
Sijin Zhou$^2$, 
Lei Meng$^1$\thanks{Corresponding author}, 
Han Hu$^3$, 
Han Yu$^4$, 
Xiangxu Meng$^1$ \\
\affiliations
$^1$School of Software, Shandong University, China \\
$^2$AIM Lab, Faculty of Engineering, Monash University, Clayton, VIC, Australia \\
$^3$School of information and Electronics, Beijing Institute of Technology, China \\
$^4$College of Computing and Data Science, Nanyang Technological University, Singapore
\emails
z\_qi@mail.sdu.edu.cn, 
sjzhou1995@gmail.com@gmail.com, 
lmeng@sdu.edu.cn,
hhu@bit.edu.cn,
han.yu@ntu.edu.sg,
mxx@sdu.edu.cn
}

\begin{document}

\maketitle

\begin{abstract}
Attribute bias in federated learning (FL) typically leads local models to optimize inconsistently due to the learning of non-causal associations, resulting degraded performance. Existing methods either use data augmentation for increasing sample diversity or knowledge distillation for learning invariant representations to address this problem. However, they lack a comprehensive analysis of the inference paths, and the interference from confounding factors limits their performance. To address these limitations, we propose the \underline{Fed}erated \underline{D}econfounding and \underline{D}ebiasing \underline{L}earning (\methodname{}) method. It constructs a structured causal graph to analyze the model inference process, and performs backdoor adjustment to eliminate confounding paths. Specifically, we design an intra-client deconfounding learning module for computer vision tasks to decouple background and objects, generating counterfactual samples that establish a connection between the background and any label, which stops the model from using the background to infer the label. Moreover, we design an inter-client debiasing learning module to construct causal prototypes to reduce the proportion of the background in prototype components. Notably, it bridges the gap between heterogeneous representations via causal prototypical regularization. Extensive experiments on 2 benchmarking datasets demonstrate that \methodname{} significantly enhances the model capability to focus on main objects in unseen data, leading to 4.5\% higher Top-1 Accuracy on average over 9 state-of-the-art existing methods.
\end{abstract}

\section{Introduction}
Federated out-of-distribution (OOD) generalization typically leverages data with diverse attributes across clients for collaborative modeling. The aim is to enhance model performance on unseen distributions \cite{qi2024attentive,qi2025global,wanghaozhao1,fu2025Alignments}. It allows FL clients to train models locally and iteratively exchange parameters with the server, enabling global aggregation without exposing private data \cite{liao2024swiss,fu2025Beyond,hu2023gitfl,liping4}. Despite its contributions to data privacy protection, existing studies show that attribute skew among sources often leads to performance degradation in FL models \cite{fu2025Beyond,zhang2024enabling,zhangopenprm,ren2025advances}. This is primarily due to spurious correlations in local models, which hinder the formation of a robust global model during aggregation \cite{yang2020federated,hu2024fedmut,cai2024fgad}.



\begin{figure}[t]
\centering
\includegraphics[width=1\linewidth]{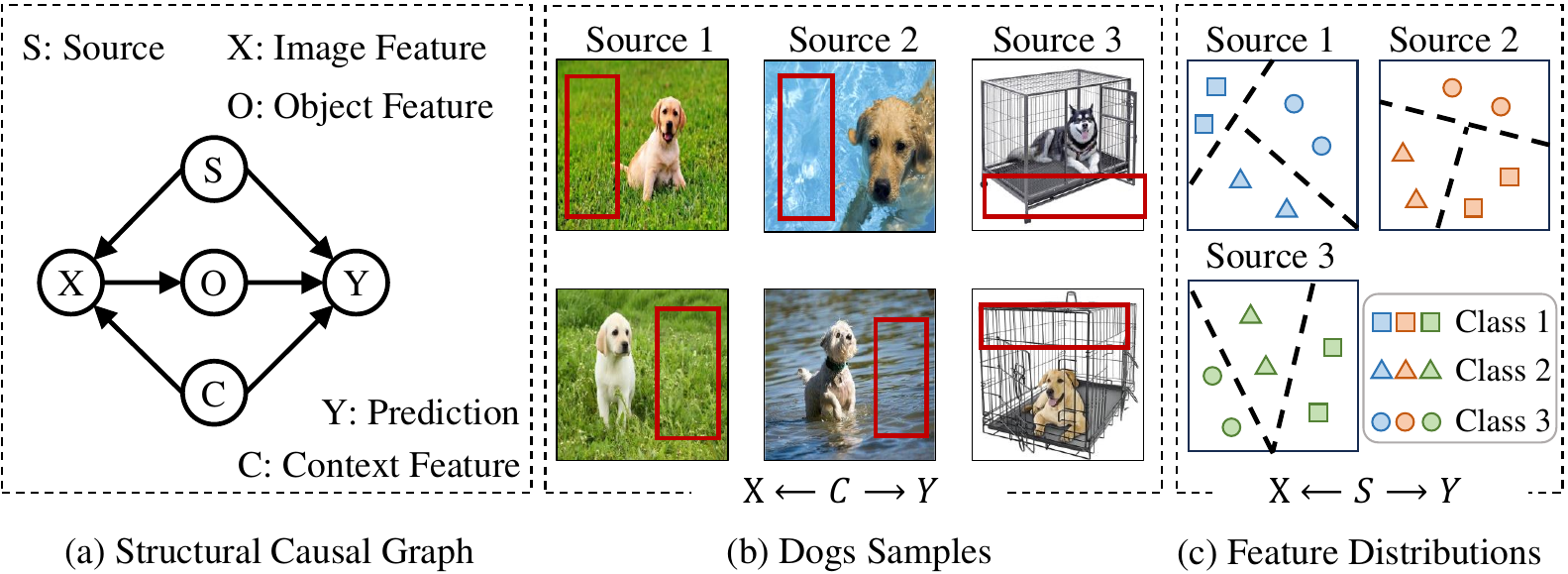}
\caption{\methodname{} reveals two factors affecting model inference: specific backgrounds and the output gap between data sources. (a) It constructs a structured causal graph to support the claim. (b) Dogs Samples illustrates the specific background within the client. (c) It represents the feature distribution differences between data sources.} 
\vspace{-0.2cm}
\label{fig1}
\end{figure}

Existing approaches for enhancing FL model performance on OOD samples can be broadly divided into two groups: 1) knowledge distillation-based methods \cite{MCGDM,fan2025ten,yu2023contrastive,huang2025Coordinator} and 2) data augmentation-based methods \cite{chen2023federated,shenaj2023learning,park2024stablefdg}. The former either treats the teacher's knowledge (e.g., the output of global or other local models) as a regularizer to provide additional supervision or decouples category-irrelevant features in the latent space, thereby mitigating the interference of irrelevant attributes. For instance, MCGDM \cite{MCGDM} shares classifiers across sources to match gradients both within and across domains to reduce overfitting to attribute features. DFL \cite{luo2022disentangled} applies mutual information-based disentanglement to address negative transfer due to attribute skew. However, the suboptimal performance of the teacher model on OOD samples and the single environment limit the effects of knowledge distillation and attribute decoupling. The latter approach typically trains generators or uses pre-trained generative models to create data with diverse attributes (e.g., CCST \cite{chen2023federated}). For example, FedCCRL \cite{wang2024fedccrl} fuses different samples for data augmentation. However, they often introduce privacy risks with performance limited by the quality of the generated data.

To address these limitation, we propose the \underline{Fed}erated \underline{D}econfounding and \underline{D}ebiasing \underline{L}earning (\methodname{}) method. As shown in Figure \ref{fig1}, it focuses on the key variables involved in the data and model inference (including data sources, background, images, objects and labels) with a structural causal model (SCM). To mitigate the joint interference of data source and background factors on model inference, we designs two main modules: 1) the intra-client deconfounding learning (DEC) module, and 2) the inter-client debiasing learning (DEB) module. Specifically, DEC decouples the background and the object in images, generating counterfactual samples to establish the association between the background and all classes without sharing any sample information among FL clients. This ensures that the model does not erroneously interpret the background as a causal factor. DEB constructs causal prototypes using object images rather than original images, which reduces the proportion of specific attributes in the prototype components. In addition, it treats causal prototypes as templates to align representations across different sources. This encourages local models to focus on target objects rather than the background, thereby promoting the learning of a unified representation space across clients.

Extensive experiments were conducted on two datasets, including performance comparisons, ablation studies and case studies with visual attention visualizations to investigate the association between background and labels. The results demonstrate that \methodname{} effectively mitigates the interference of background and focuses on the objects of samples in unseen cases, improving the effectiveness of collaborative learning. Compared to nine state-of-the-art existing methods, \methodname{} achieves 4.5\% higher Top-1 Accuracy on average.


\section{Related Work}

\subsection{Data Augmentation-based Methods}
To mitigate attribute bias, data augmentation-based methods~\cite{wang2024fedccrl,shenaj2023learning,liu2021feddg,xu2023federated,zhang2024federated,de2022mitigating,zhang2025federated,qi2022clustering} have been developed to improve model generalization to previously unseen attributes by boosting the diversity of the data attributes. These methods typically rely on two approaches: 1) training data generators to create novel samples, or 2) utilizing pre-trained diffusion models to enrich sample diversity. The first approach involves exchanging local information between clients to generate new samples from different domains. For instance, FIST~\cite{nguyen2024fisc} and StableFDG~\cite{park2024stablefdg} produce samples with varying styles by sharing style-related information across clients. The second approach leverages prompt-driven techniques to generate samples that align with certain criteria \cite{morafah2024stable,zhao2023federated}. Although enhancing diversity, these strategies often introduce privacy concerns. Moreover, the quality differences between generated and original samples also limit their effectiveness.

\begin{figure}[!b]
\centering
\includegraphics[width=1\linewidth]{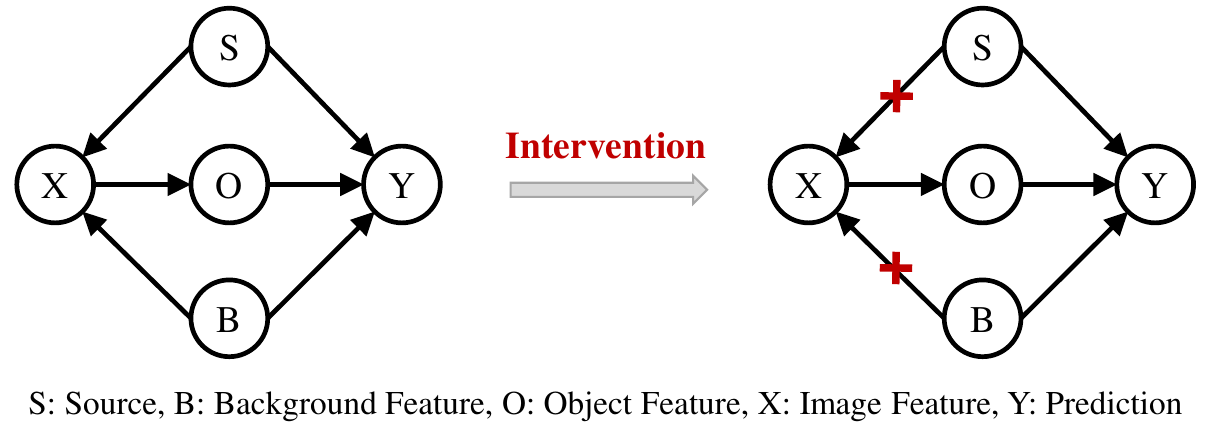}
\caption{A causal view in federated out-of-distribution generalization, which uses backdoor intervention to eliminate the interference of background factor \(B\) and data source factor \(S\) on the sample \(X\) during model inference (i.e., $B\nrightarrow X$ and $S\nrightarrow X$).
} 
\label{fig2}
\end{figure}

\begin{figure*}[!t]
\centering
\includegraphics[width=1\linewidth]{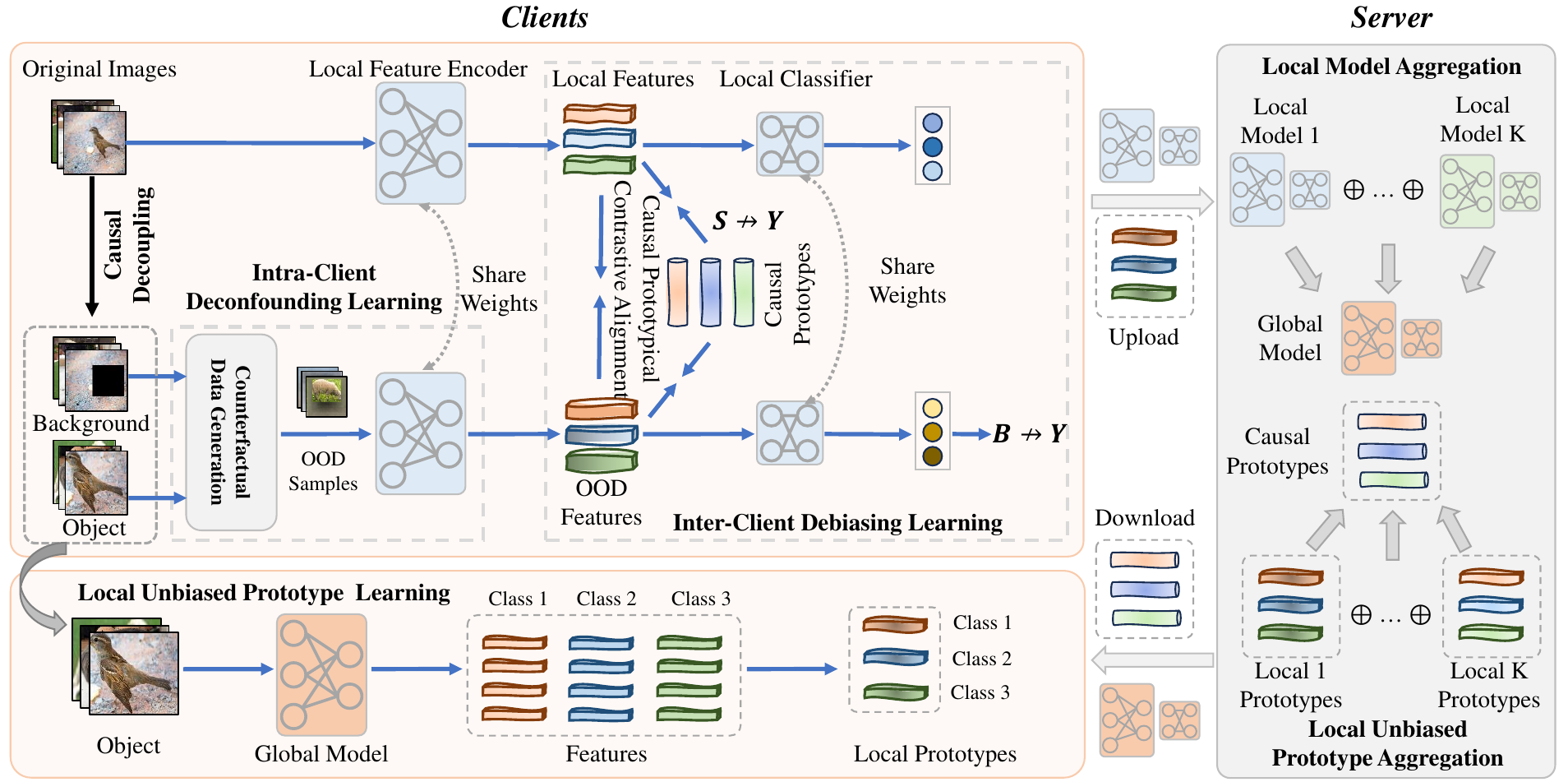}
\caption{The \methodname{} framework. It contains two main modules: 1) the intra-client deconfounding learning module, and 2) the inter-client debiasing learning module. The former generates counterfactual samples to break the spurious association between the background and specific labels $B\nrightarrow X$. The latter leverages causal prototypes to promote consistency in learning heterogeneous representations $S\nrightarrow X$.
} 
\label{fig3}
\end{figure*}
\subsection{Knowledge Distillation-based Methods}
Knowledge distillation-based approaches utilize generalized knowledge to guide local models in extracting shared features that are independent of specific attributes. These methods typically adopt two main strategies: 1) regularizing representations from different sources to align them, thus promoting consistency in representations across diverse contexts~\cite{wang2023dafkd,qi2024crosstraining,qi2025cross,qi2023cross,liu2021feddg,sun2023feature,meng2024improving,zhu2024dualfed}; or 2) decoupling invariant features within the latent space to minimize the impact of confounding factors~\cite{yu2023contrastive,liping2,yu2011dynamic,cai2024lgfgad}. For instance, FedProc~\cite{mu2023fedproc} and FPL~\cite{fpl} leverage prototypes to enforce consistent representation learning across all clients, thereby encouraging them to converge within a shared representation space. MCGDM~\cite{MCGDM} mitigates overfitting within individual domains by employing both intra-domain and cross-domain gradient matching to improve generalization. Similarly, FedIIR~\cite{fediir} enables the model to implicitly learn invariant relationships by capitalizing on prediction inconsistencies and gradient alignment across clients. Although these techniques have achieved notable performance improvements, they often rely on single-domain data, which hinders model transferability to OOD domains. Our \methodname{} method bridges these gaps.

\section{Preliminaries}
An FL system typically involves multiple distinct clients, denoted as \( \mathcal{S} = \{S_1, S_2, \dots, S_K\} \), and a central coordinating server \( C \). Each \( S_k \) utilizes its own private dataset \( D^k = \{(X^k, Y^k)\} \) to train a local model \( M_k \). The server \( C \) aggregates the local model parameters \( \{\theta_k\} \) from all participating clients to compute the global parameters \( \theta_g = \sum_{k=1}^{K} \alpha_k \theta_k \), where \( \alpha_k \) is the weight based on the size of each client's local dataset, and \( \alpha_k = \frac{|D^k|}{\sum_{k=1}^{K} |D^k|} \).

Under this setting, \methodname{} constructs a structural causal graph to analyze the confounding factors involved in FL model inference. As shown in Figure \ref{fig2}, the main idea is to tackle the interference of client data heterogeneity ($S \rightarrow X $) and background ($B \rightarrow X$) factors in model decision-making, where $X$ is the input image. The intra-client deconfounding learning module to generate counterfactual data to improve data diversity, which can sever the link between background and specific class during inference (i.e., $B\nrightarrow  X$). The inter-client debiasing module constructs class-wise causal prototypes $U_c = {u_c^1, \dots, u_c^N}$ from object images, and uses them to align heterogeneous representations across clients. This mitigates the influence of client-specific model differences during aggregation (i.e., $S \nrightarrow X$).


\section{The Proposed \methodname{} Method}
This section presents the proposed \methodname{} method with reference to the schematic framework shown in Figure \ref{fig3}. 

\subsection{A Causal View on FL OOD Generalization}\label{sec4.2}
To clarify the factors that influence model inference in federated OOD generalization, we have constructed a structured causal graph that comprehensively illustrates the potential inference paths of the model, as shown in Figure \ref{fig2}. Specifically,
\begin{itemize}[leftmargin=10pt]
    \item $X\leftarrow B \rightarrow Y$ indicates that the image background \( B \) is a potential confounding factor, causing the model to rely on it when inferring Label \( Y \) for Sample \( X \). The proposed intra-client deconfounding learning module is designed to sever the connection between \( x \) and \( B \) (i.e., $X\nleftarrow B$).

    \item $X\leftarrow S \rightarrow Y$ implies that different clients contain diverse image features. Bridging this heterogeneity can eliminate interference between images and their labels. The proposed inter-client debiasing learning module is designed to align heterogeneous representations, which severs the association between $S$ and input image $X$ (i.e., $X\nleftarrow S$).

    \item $X\rightarrow O\rightarrow Y$ means that the design of all modules focuses on learning the true causal effects, which infers labels $Y$ based on key objects $O$ in an image $X$, while avoiding spurious correlations with confounding factors $B$ and $S$. By severing both paths, the causal structure is clearly revealed during federated OOD generalization.

\end{itemize}

\subsection{Intra-Client Deconfounding Learning (DEC)}
To address shortcut learning caused by client-specific backgrounds, the DEC module aims to block the causal link from background to label during inference, encouraging the model to focus on true causal relationships.
To achieve this, it performs backdoor intervention to adjust the sample distribution $do(X)$. By applying the law of total probability, the inference involves both the direct causal impact of $X$ on $Y$ and the correlation confounded by $B$:
\begin{equation}
\resizebox{0.65\hsize}{!}{$
   P(Y \mid X)= {\textstyle \sum_{B}^{}}  P(Y \mid X, B) P(B \mid X)$}.
\end{equation}

To ensure that each background factor contributes equally to label inference, the DEC module generates counterfactual samples to intervene in the distribution:
\begin{equation}
\begin{aligned}
    P(Y \mid do(X)) &= \textstyle \sum_{B} P(Y \mid DEC(X, B)) \\
                    &= \textstyle \sum_{B} P(Y \mid X, B) P(B),
\end{aligned}
\end{equation}
which removes the path from \(B\) to \(X\) in Figure \ref{fig2}, thereby facilitating the model to approximate the causal intervention \(P(Y \mid \text{do}(X))\) instead of the spurious correlation \(P(Y \mid X)\).
Specifically, the DEC module begins by decoupling images into objects and backgrounds. Then, it modifies the background factors associated with the objects. Inspired by the pre-trained model Grounding DINO \cite{liu2025grounding}, the approach leverages text prompts $T$ to detect targets:
\begin{equation}
\resizebox{0.47\hsize}{!}{$
  \theta= \begin{bmatrix}
  \theta_1 & \theta_2 \\
  \theta_3 & \theta_4
\end{bmatrix} = DINO(X, T),$}
\end{equation}
\begin{equation}\resizebox{0.91\hsize}{!}{$
I_{O}=I \odot 1_{(x, y) \in\left[\theta_{1}, \theta_{2}\right] \times\left[\theta_{3}, \theta_{4}\right]}, I_{B}=I \odot 1_{(x, y) \notin\left[\theta_{1}, \theta_{2}\right] \times\left[\theta_{3}, \theta_{4}\right]},$}
\end{equation}
where \( I_O \) and \( I_B \) are the object and background images. \( T \) is the label name. \( \theta \) represents the bounding box coordinates (\( \theta_1, \theta_3 \) for the top-left, and \( \theta_2, \theta_4 \) for the bottom-right). \( \odot \) denotes the Hadamard product. \( 1_{(x, y) \notin [x_1, x_2] \times [y_1, y_2]} \) is an indicator function, which evaluates to 1 outside the bounding box and 0 inside it. 

\begin{figure}[t]
\centering
\includegraphics[width=1\linewidth]{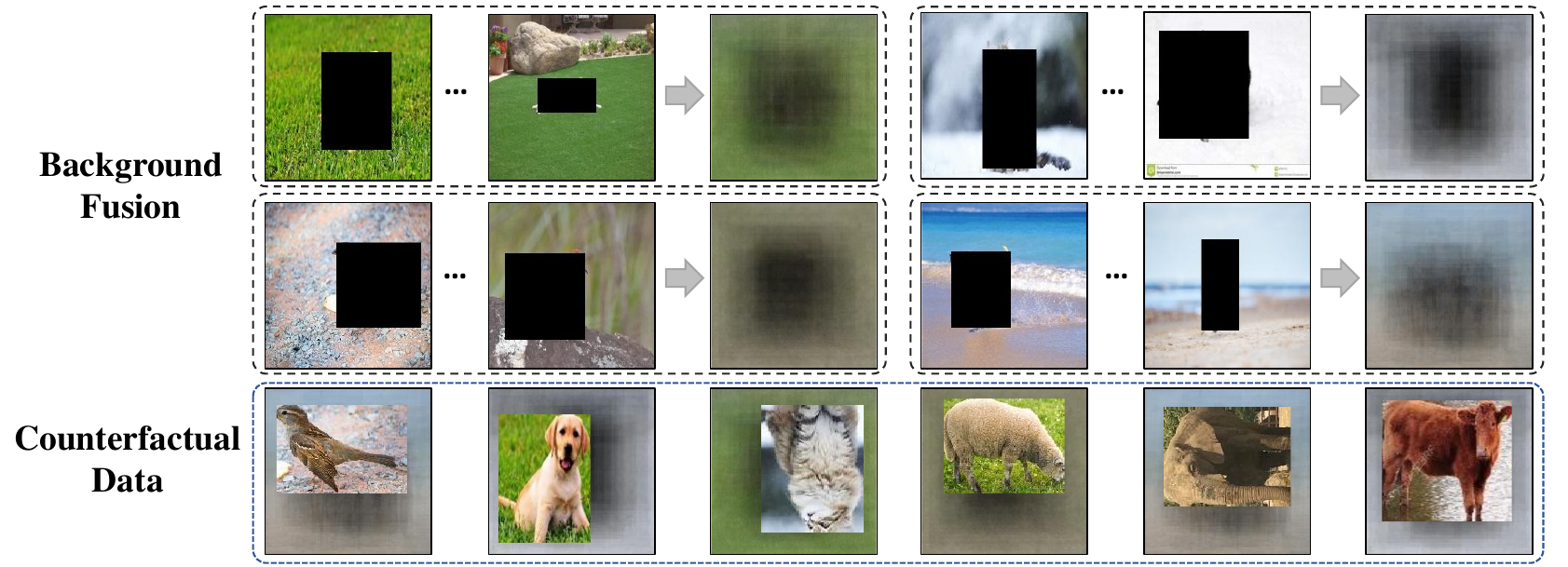}
\caption{Example counterfactual samples with causal features.
} 
\label{fig4}
\end{figure}

Subsequently, to generate counterfactual samples with causal features and diverse backgrounds for distribution intervention, the DEC module divides the backgrounds $I_B^i$ belonging to class $i$ into $\eta$ groups, denoted as $\{g_i^j|j=1,\dots,\eta\}$:
\begin{equation}\{g_i^j|j=1,\dots,\eta\}=Split(I_B^i, \eta),
\end{equation}
where $Split(\cdot)$ is the random grouping function. The counterfactual data $I_C$ can be created by fusing randomly grouped backgrounds $g_i^j$ and an object $I_O$:
\begin{equation}
\resizebox{0.65\hsize}{!}{$
    I_C= \left [ \frac{1}{|g_i^j|} {\textstyle \sum_{I_B\in g_i^j}^{}} I_B \right ] \oplus Trans(I_O),$}
\end{equation}
where $Trans(\cdot)$ denotes a transformation function, which may include rotation, flipping and resizing, as shown in Figure \ref{fig4}.
To guide the model to focus on the target objects in different cases, we optimize it using a joint classification loss:
\begin{equation}
\resizebox{0.91\hsize}{!}{$
\mathcal{L}_{J}=- \sum_{n=1}^{N} y(n) \log \left( M_k(I)(n) \right)+- \sum_{n=1}^{C} y_C(n) \log \left( M_k(I_C)(n) \right),$}
\end{equation}
where $N$ is the total number of classes. $M_k(\cdot)$ is the model of client $k$. $y(n)$ and $y_c(n)$ are the true labels of $n-th$ index for the original and counterfactual samples. $M_k(I)(n)$ and $M_k(I_C)(n)$ are corresponding predictions.

\begin{algorithm}[!t]
\caption{\textsc{FedDDL}}
\begin{algorithmic}[1]
\State Initialize the global model parameter $\theta^0$
\For{$t = 1, \dots, T$}
    \State Sample subset $\mathcal{K}$ of clients with $|\mathcal{K}| = k$
    \For{each client $k \in \mathcal{K}$ in parallel}
        \State Initialize local model parameter $\theta_k^t = \theta^{t-1}$
        \State Counterfactual sample generation $D_{C,k}$ based on objects $I_{O,k}$ and backgrounds $I_{B,k}$ images
        \For{$e = 1, \dots, E$}
            \State Sample batches of data $\zeta_1$, $\zeta_2$ from local data $D^k$ and counterfactual data $D_{C,k}$ 
            \If {$t$ = 1}
            \State $g_k = \nabla \mathcal{L}_J(\zeta_1, \zeta_2)$
            \Else
            \State $g_k =\nabla \mathcal{L}_J(\zeta_1, \zeta_2)+\lambda \nabla \mathcal{L}_{CR}(\zeta_1, \zeta_2, U_G)$
            \EndIf
            \State Update $\theta_k^t \gets \theta_k^t - \eta_l g_k$
        \EndFor
         \State $U_{L,t}^{i,k}\gets\frac{1}{|I_O^{i,k}|} {\textstyle \sum}  M_G(\theta^{t-1},I_O^{i,k})$, $i = 1, \dots, N$
    \EndFor
    \State $\theta^t = 1/k \sum_{k \in \mathcal{K}} \theta_k^{t-1}$
    \State $U_{G,t}^i = 1/k \sum_{k \in \mathcal{K}} U_{L,t}^{i,k}$, $i=1,\dots, N$
    
\EndFor
\end{algorithmic}\label{alg1}
\end{algorithm}

\subsection{Inter-Client Debiasing Learning (DEB)}
To address the issue of conflicts among clients leading to declines in the overall FL model performance, the DEB module aims to sever the path from FL clients to labels in order to arrive at consistent decisions for the same input.
It performs backdoor adjustments to the representation learning of each client. The original multi-source inference is expressed as:
\begin{equation}
\resizebox{0.91\hsize}{!}{$
P(Y | X) = \sum_{i=1}^{K} \sum_{B} P(Y | X, B, S_i) P(B | X, S_i) P(S_i | X),$}
\end{equation}
which includes both the direct causal effect of \(X\) on \(Y\) and the correlation confounded by \(B\) and \(S\). The DEC module mitigates the interference from factor \(B\) by severing the connection between the data source \(S\) and the label \(Y\):
\begin{equation}
\resizebox{0.91\hsize}{!}{$
\begin{aligned}
P(Y | do(X)) & = \textstyle \sum_{i=1}^{k}  \sum_{B}^{} P(Y | DEC(X, B), DEB(S_i)) \\
 & = \textstyle \sum_{i=1}^{k}  \sum_{B}^{} P(Y | X, B, S_i) P(B) P(S_i),
\end{aligned}$}
\end{equation}
Specifically, it first constructs causal prototypes using the extracted object images $I_O$ to reduce the proportion of background features in the prototype components as:
\begin{equation}
    U_L^{i,k}=\frac{1}{|I_O^{i,k}|} {\textstyle \sum_{\hat{I}\in I_O^{i,k} }^{}}  M_G(\hat{I}),
\end{equation}
where $U_L^{i,k}$ and $I_O^{i,k}$ represent the local causal prototype of class $i$ and the data set in client $k$, respectively.
$M_G(\cdot)$ denotes the global model.  Moreover, the global causal prototype $U_G^{i,k}$ of class $i$ can be represented as:
\begin{equation}
    U_G^{i}=\frac{1}{K} {\textstyle \sum_{k=1}^{K}}  U_L^{i,k}.
\end{equation}

\begin{figure}[!t]
\centering
\includegraphics[width=1\linewidth]{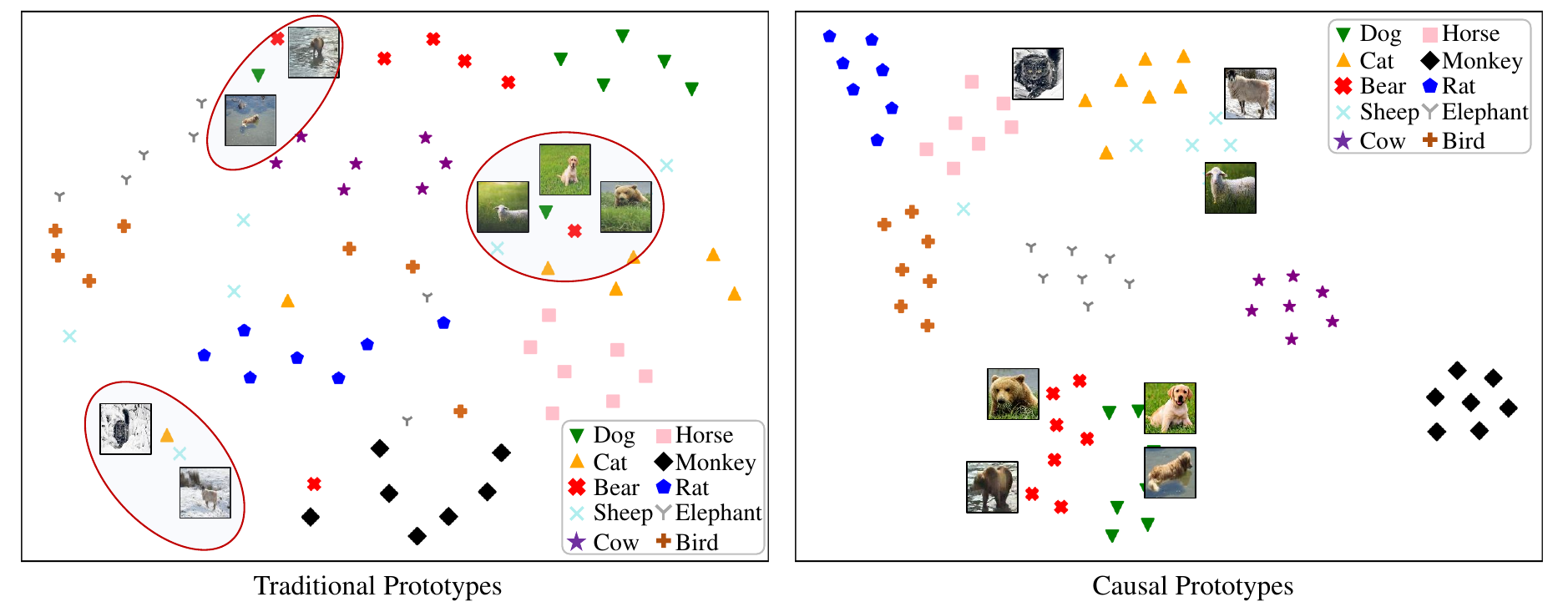}
\caption{Comparison between traditional and causal prototypes. Causal prototypes alleviate the interference from specific attributes.} 
\label{fig5}
\end{figure}

As shown in Figure \ref{fig5}, backgrounds might cause different class prototypes to exhibit similar distributions across FL clients. In contrast, causal prototypes effectively eliminate the interference of backgrounds, thereby enhancing their distinction. Therefore, the DEB module performs causal prototype-based regularization to guide the adjustment of representation distributions in each client:
\begin{equation}
\resizebox{0.85\hsize}{!}{$
\mathcal{L}_{CR}  =  - \log \frac{{\exp (f \cdot U_G^ +  /\tau )}}{{\exp (f \cdot U_G^ +  /\tau ) + \sum {\exp (f \cdot U_G^ -  /\tau )} }}$},
\end{equation}
where $f$ represents a local feature. $U_G^+$ and $U_G^-$ denote the global causal prototypes of the same and different classes as $f$, respectively. $\tau$ is the temperature parameter. This reduces the output gap between models from different clients, and eliminates interference from clients' local factors.

\subsection{Training Strategy}
\methodname{} aims to eliminate the interference of client-specific attributes and decision differences on the final model inference. This is achieved through the collaboration of the DEC module and the DEB module. We summarize the pseudo-code of \methodname{} in Algorithm \ref{alg1}. Its overall optimization objective can be express as:
\begin{equation}
    \mathcal{L}_{total} = E_{(x,y)\sim D_{local}}[\mathcal{L}_{J} + \lambda *  \mathcal{L}_{CR}],
\end{equation}
where $\lambda$ is a weighted parameter.

\section{Experimental Evaluation}
\subsection{Experiment Settings}
\paragraph{Datasets}
Following previous work on OOD generalization \cite{qi2024attentive,wang2022meta}, experiments are conducted on two datasets: NICO-Animal and NICO-Vehicle~\cite{NICO_Causal}. Their statistics and partitioning method of the datasets can be found in Table \ref{tab1}.

\begin{table}[ht]
    \centering
    \begin{tabular}{c|c|c|c } 
    \hline
     \textbf{Datasets} & \textbf{\#Class} & \textbf{\#Training} & \textbf{\#Testing} \\ 
    \hline
    NICO-Animal (F7) & 10 & 10,633 & 2,443 \\
    \hline
    NICO-Animal (L7) & 10 & 8,311 & 4,765 \\
    \hline
    NICO-Vehicle (F7) & 10 & 8,027 & 3,626 \\
    \hline
    NICO-Vehicle (L7) & 10 & 8,352 & 3,301 \\
    \hline
  \end{tabular}
    \caption{Statistics of NICO-Animal and NICO-Vehicle. F7 represents data from the first seven backgrounds of each class used as the training set. L7 represents data from the last seven backgrounds of each class used as the training set.}\label{tab1}
\end{table}

\paragraph{Evaluation Metrics}
Following prior work~\cite{fedavg,li2021model}, we use Top-1 Accuracy to evaluate performance, which is defined as \( \frac{1}{N} \sum_{i=1}^{N} \mathbb{I}(\hat{y}_i = y_i) \), where \( N \) is the total number of samples, \( \hat{y}_i \) is the predicted label, and \( y_i \) is the ground-truth label.


\paragraph{Network Architecture}
To ensure fair comparison, all methods adopt the same network architecture. In \methodname{}, a unified architecture is used to process both raw images and those processed with Grounding DINO. Following prior studies~\cite{moon,liu2021feddg}, ResNet-18~\cite{he2016deep} is selected as the backbone for both datasets.

\paragraph{Implementation Details}
We set the local training epochs to 10 per global round for both datasets. The total number of communication rounds is 50, with 7 clients for both datasets. We used a client sampling fraction of 1.0 and employed SGD as the optimizer. During local training, the weight decay is set to 0.01, the batch size is 64, and the initial learning rate is 0.01 for both datasets. $\lambda$ is chosen from the set $\{0.1, 1.0, 2.0\}$. $\tau$ is selected from $\{0.5, 0.07\}$. $\eta$ is tuned from $\{1,3,5\}$. The BOX\_THRESHOLD and TEXT\_THRESHOLD in the DINO model have been set to 0.3. For other methods, hyperparameters are tuned based on the corresponding papers.

\subsection{Performance Comparison}
This section presents a comparison of the \methodname{} method with nine state-of-the-art (SOTA) methods, including FedAvg \cite{fedavg}, FedProx \cite{fedprox}, MOON \cite{li2021model}, FPL \cite{fpl}, FedIIR \cite{fediir}, FedDecorr \cite{10336535}, FedHeal \cite{Fedheal}, MCGDM \cite{MCGDM}, and FedCCRL \cite{wang2024fedccrl}. The results derived from Table \ref{tab:comparison} are summarized below.

\begin{itemize}[leftmargin=10pt]
\item \methodname{} achieve significant improvements across all cases compared to existing methods. This includes evaluations of both global (Global) and local (Local) models, where the consistent enhancements highlight the robustness of \methodname{} in coping with complex scenarios.

\item  The DEC module is a plug-and-play component that can be easily integrated into other methods, such as MOON, and can provide significant performance improvements. This is reasonable because it enhances sample diversity and ensures the preservation of causal features.
\item  By addressing the federated OOD problem from both intra-client and inter-client perspectives, \methodname{} outperforms single-perspective methods like MOON and FPL. \methodname{} leverages deconfounding learning to disrupt spurious correlations between backgrounds and labels within clients, while also employing debiasing learning to reduce the output gap between clients. 
\item \methodname{} also reduces the performance disparity between different local models (as reflected by the local variance). This can be understood as it simultaneously strengthens the performance of individual local models via the DEC module while leveraging the DEB module to promote consistency in outputs across heterogeneous clients. 

\end{itemize}

\begin{table*}[t]
    \centering
    \resizebox{1\textwidth}{!}{
    \setlength{\tabcolsep}{3mm}{
    \begin{tabular}{c|c|c|c|c|c|c|c|c}
    \hline   
    \multicolumn{1}{c|}{\multirow{3}*{\textbf{Methods}}} & \multicolumn{4}{c|}{\textbf{NICO-Animal}} & \multicolumn{4}{c}{\textbf{NICO-Vehicle}} \\
    \cline{2-9}
    &\multicolumn{2}{c|}{\textbf{Global}} &\multicolumn{2}{c|}{\textbf{Local}} &\multicolumn{2}{c|}{\textbf{Global}} &\multicolumn{2}{c}{\textbf{Local}} \\\cline{2-9}
    \multicolumn{1}{c|}{} & F7 & L7 & F7 & L7 & F7 & L7 & F7 & L7 \\
    \hline
    FedAvg & 44.38±0.6 & 52.75±0.6 & 34.19±4.6 & 44.45±4.6 & 63.28±0.4 & 59.05±0.2 & 48.44±6.9 & 44.79±6.2 \\
    Fedprox & 44.55±0.9 & 51.99±0.9 & 35.72±4.9 & 43.17±5.3 & 65.36±0.6 & 57.50±0.8 & 47.62±5.1 & 42.63±5.3 \\
    MOON & 45.53±0.4 & 53.66±0.9 & 36.31±5.7 & 46.12±6.2 & 65.94±0.5 & 59.63±0.5 & 51.34±5.8 & 46.54±5.2 \\
    FPL & 47.76±0.5 & 55.39±0.2 & 37.58±4.5 & 46.93±4.4 & 68.51±0.7 & 61.76±0.6 & 52.22±5.1 & 48.26±4.9 \\
    FedDeccor & 48.11±0.6 & 53.12±0.2 & 37.64±5.2 & 47.32±5.1 & 67.39±0.7 & 61.32±0.6 & 51.21±6.7 & 46.34±5.3 \\
    FedIIR & 46.40±0.9 & 52.82±0.7 & 35.71±4.9 & 45.88±4.7 & 63.64±0.9 & 56.18±0.4 & 49.47±5.1 & 47.83±5.4 \\
    FedHeal & 42.32±1.0 & 52.80±0.6 & 36.44±5.1 & 44.78±3.6 & 64.00±0.5 & 56.25±0.7 & 46.32±5.2 & 44.59±4.9 \\
    MCGDM & 47.96±0.8 & 54.53±0.5 & 38.46±4.8 & 47.51±3.5 & 66.84±0.4 & 59.59±0.9 & 53.69±7.9 & 46.92±4.4 \\
    FedCCRL & 48.49±0.7 & 57.31±0.9 & 39.81±3.9 & 47.37±4.2 & 70.14±0.5 & 62.23±0.9 & 54.73±6.8 & 50.32±7.6 \\\hline
    MOON+DEC & 52.47±0.9 & 60.13±0.7 & 42.56±5.9 & 50.26±6.1 & 71.43±0.6 & 64.45±1.1 & 56.32±4.4 & 52.56±5.7 \\
    \methodname{} & \textbf{53.37±0.4} & \textbf{62.59±0.6} & \textbf{44.27±3.1} & \textbf{53.23±3.0} & \textbf{73.38±0.7} & \textbf{66.20±0.4} & \textbf{59.51±4.6} & \textbf{54.28±4.2} \\
    \hline
    \end{tabular}}}
    \caption{Performance comparison between \methodname{} and baselines on NICO-Animal and NICO-Vehicle datasets. All methods were executed across three trials, and the results are reported as the mean and standard deviation of the top-1 accuracy.}    
    \label{tab:comparison}
\end{table*}

\begin{table}[t]
    \centering
     \resizebox{1\linewidth}{!}{
    \begin{tabular}{c|c|c|c|c}
         \hline
         \multicolumn{1}{c|}{\multirow{2}*{}} & \multicolumn{2}{c|}{\textbf{NICO-Animal (L7)}} & \multicolumn{2}{c}{\textbf{NICO-Vehicle (L7)}} \\
         \cline{2-5}
         \multicolumn{1}{c|}{}&Global &Local &Global &Local\\
        \hline
         \textbf{FedAvg}&52.75±0.6&44.45±4.6&59.05±0.2&44.79±4.2\\
        \hline
        \textbf{+ $\mathrm{DEB}_{\mathrm{L}}$ ${}$}&54.39±0.5&48.74±4.1&62.43±0.7&47.82±3.6\\
        \textbf{+ $\mathrm{DEB}_{\mathrm{C}}$ ${}$}&55.12±0.8&49.13±3.2&62.58±0.4&49.37±3.1\\
       \textbf{+ $\mathrm{DEB}_{\mathrm{L+C}}$ ${}$}&57.43±0.7&49.56±3.4&63.49±0.6&50.74±3.3\\
        \textbf{ + $\mathrm{DEC}$ ${\mathrm{}}$}&57.78±0.6&48.63±2.4&63.11±0.5&51.36±2.7\\
        \textbf{ + $\mathrm{DEB}_{\mathrm{L}}$ + $\mathrm{DEC}_{\mathrm{}}$}&60.84±0.4&51.68±2.1&64.15±0.3&52.37±2.4\\
        \textbf{ + $\mathrm{DEB}_{\mathrm{C}}$ + $\mathrm{DEC}_{\mathrm{}}$}&61.69±0.7&52.31±2.7&65.28±0.4&53.24±2.7\\
        \textbf{+ $\mathrm{DEB}_{\mathrm{L+C}}$ + $\mathrm{DEC}_{\mathrm{}}$}&\textbf{62.54±0.5}&\textbf{53.23±2.7}&\textbf{66.20±0.4}&\textbf{54.28±2.2}\\
        
         \hline
    \end{tabular}}
     \caption{Ablation study of \methodname{} on NICO-Animal (L7) and NICO-Vehicle (L7) with top-1 accuracy.}    
    \label{tab:ablation}
\end{table}

\subsection{Ablation Study}
This section provides an in-depth analysis of the effectiveness of various modules within the \methodname{}, including the intra-client deconfounding learning ($DEC$) module, the inter-client decbiasing learning module with local data ($DEB_{L}$), counterfactual data ($DEB_{C}$), and their combined set ($DEB_{L+C}$). The results are presented in Table \ref{tab:ablation}.

\begin{itemize}[leftmargin=10pt]
\item  Incorporating the inter-client decbiasing learning module contributes to improving the performance of the global model and the average performance of the local models, but there is a significant variation in the performance across different client models.

\item The inter-client decbiasing learning module plays a significant role in bridging the performance gap between client models, which promotes consistent improvements across heterogeneous sources and enhances the alignment of model representations, diminishing discrepancies in client-specific performance.

\item As the diversity of the data increases, the advantages of the DEB module are amplified, as it effectively enhances heterogeneous representation alignment, particularly in complex cases, where its ability to adapt and align diverse representations proves to be a significant asset.


\end{itemize}

\begin{figure}[t]
\centering
\includegraphics[width=1\linewidth]{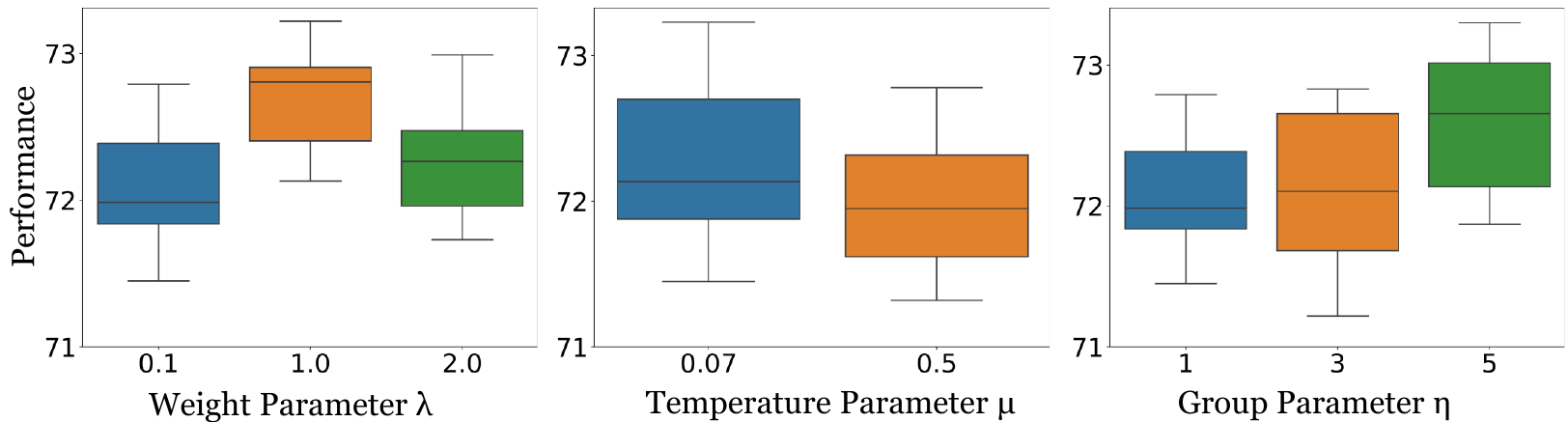}
\caption{The impact of hyperparameters on performance. The $\lambda$, $\tau$ and $\eta$ are tuned from $\{0.1,1.0,2.0\}$, $\{0.07,0.5\}$ and $\{1,3,5\}$ on NICO\_Vehicle (F7), respectively.
} 
\label{fig6}
\end{figure}

\subsection{Robustness of \methodname{} on Hyperparameters}
This section assesses the robustness of \methodname{} under different hyperparameters. Specifically, we explore the impact of hyperparameters $\lambda$, $\tau$, and $\eta$ by selecting values from the sets $\{0.1,1.0,2.0\}$, $\{0.07,0.5\}$, and $\{1,3,5\}$, respectively. As depicted in Figure \ref{fig6}, \methodname{} consistently outperforms FedAvg across various scenarios, showing remarkable insensitivity to hyperparameter variations within a broad range. This suggests that \methodname{} is highly robust and adaptable to changes in hyperparameters. For the adjustment of $\tau$, setting $\tau = 0.07$ enhances model performance compared to $\tau = 0.5$ by increasing the representation differences between sources. This adjustment allows the model to focus more on these variations, leading to improved generalization across clients. For $\eta$, increasing \(\eta\) means greater data diversity. The greater the diversity in the data, the more the model benefits, highlighting the importance of diverse data sources in improving the model's ability to generalize.

\subsection{Case Study}

\subsubsection{Breaking the Background-Category Association}
\begin{figure}[t]
\centering
\includegraphics[width=1.0\linewidth]{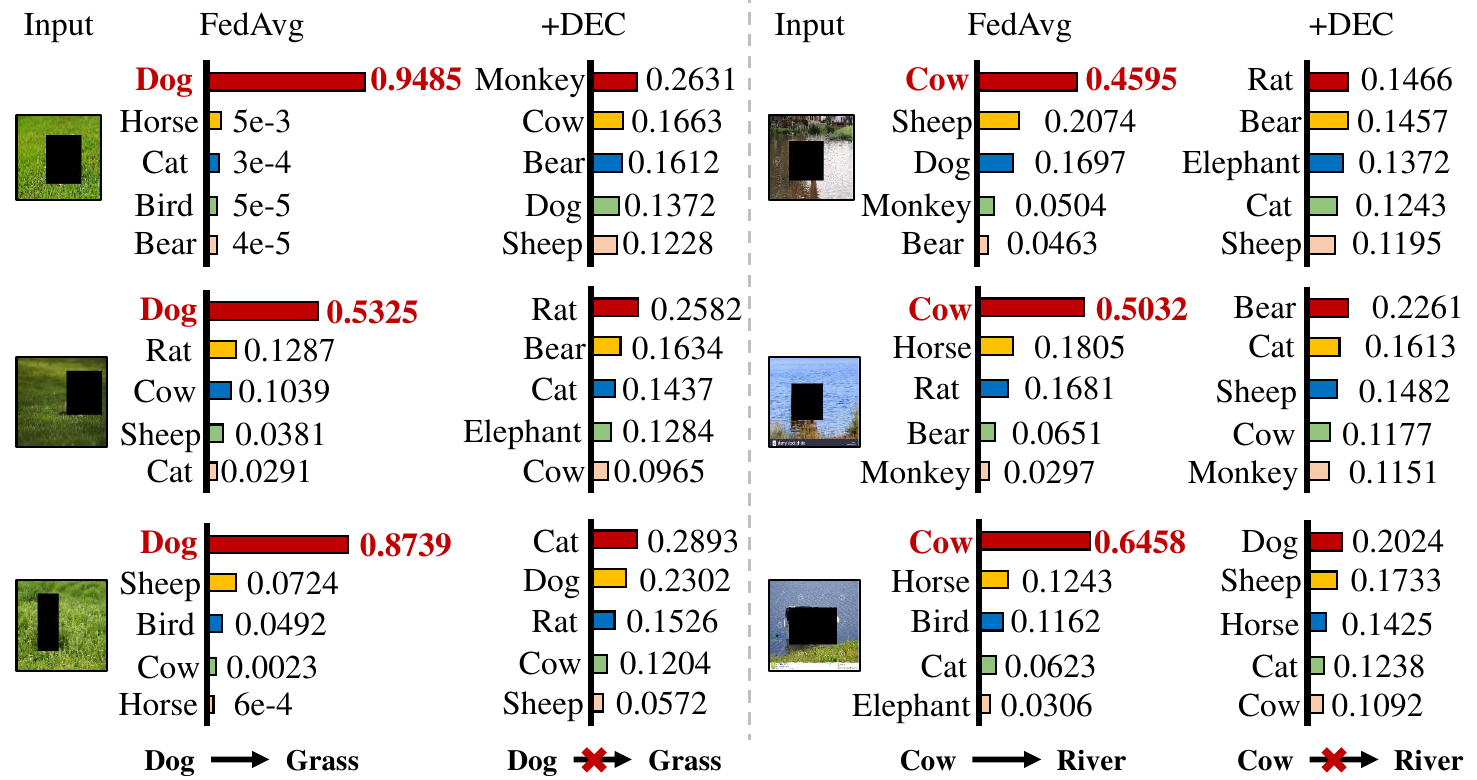}
\caption{(a) FedAvg forms spurious associations between ``dog" and ``grass", as well as ``cow" and ``river"; (b) The DEC module helps the model reduce the probability gap between the background being predicted as any category.
} 
\label{fig7}
\end{figure}

Here, we examine the effectiveness of the DEC module in breaking the association between background and specific labels. As shown in Figure \ref{fig7}(a), the unique attributes of the data lead to spurious associations in the FedAvg method between ``dog" and ``grass" as well as ``cow" and ``river". Even when the objects in the images are masked, the model still identifies the background with an exceptionally high confidence as the label of the object. This can easily cause the model to rely on non-causal associations to infer the labels of samples, ultimately leading to errors. Moreover, benefiting from the intervention of the DEC module, the association between background and labels is disrupted, leading the model to predict the background with equal probability across all dimensions. As shown in Figure \ref{fig7}(b), it narrows the gap among the probabilities of predicting the background as different categories, avoiding to use a very robust but causally wrong feature to make predictions.


\begin{figure}[h]
\centering
\includegraphics[width=1.0\linewidth]{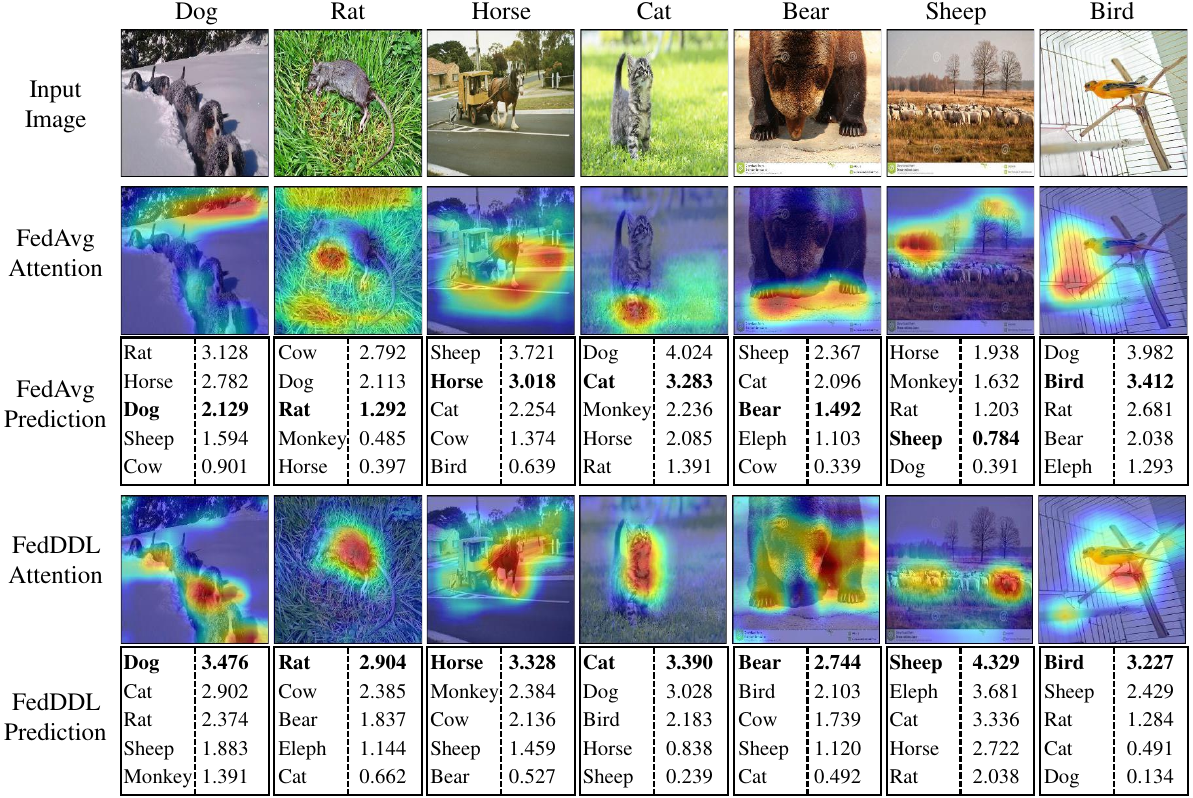}
\caption{Visualization of visual attention and comparison of predictions for FedAvg and \methodname{}. \methodname{} enhances the model's ability to focus on target objects in out-of-distribution samples and improves the confidence in ground-truth predictions.
} 
\label{fig8}
\end{figure}
\subsubsection{Comparison of Visual Attention}
This section analyzes the effectiveness of the deconfounding learning in local 1. We randomly selected test samples and visualized the visual attention of different methods using GradCAM \cite{selvaraju2017grad,slazyk2022cxr,lin2020multi,meng2019learning}. As illustrated in Figure \ref{fig8}, FedAvg often focuses on the background of out-of-distribution samples, leading to classification errors due to its limited generalization ability. For example, in the case of a cat on grass, FedAvg may assign the label based on ``grass," leading to incorrect identification as a ``dog," since the training data only contains dogs on grass. Furthermore, \methodname{} effectively focuses on the target object in test samples, enabling the model to eliminate background interference and make accurate predictions. This significantly enhances the generalization ability of local models and demonstrates that improving the performance of individual models also contributes to better collaboration.

\subsubsection{Analysis of Cross-Silo Representation Alignment}

\begin{figure}[t]
\centering
\includegraphics[width=1.0\linewidth]{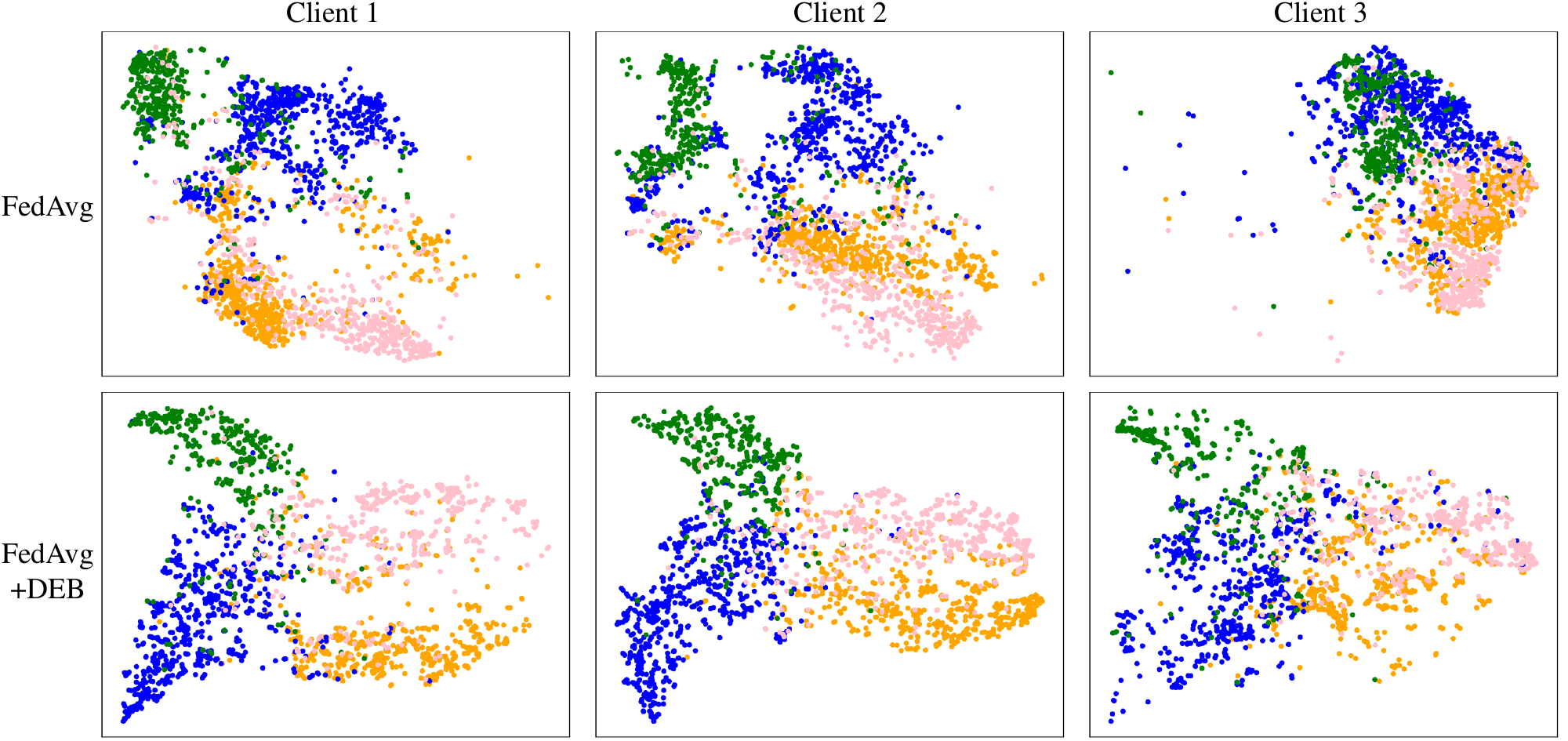}
\caption{Visualization of representation distributions from different sources. The DEB module has made significant contributions to cross-silo feature alignment in out-of-distribution cases.
} 
\label{fig9}
\end{figure}

This section analyzes the consistency of representation distributions from models of different sources on the same test samples. As shown in Figure \ref{fig9}, the image representations from clients 1, 2 and 3 are visualized using t-SNE \cite{van2008visualizing,zhou2023fedfa}. Clearly, the DEB module has mitigated the gap in outputs from models of different sources. Meanwhile, it has also enhanced the discriminability of representations across classes. This is reasonable because causal prototype contrastive alignment helps alleviate overfitting to specific attributes and facilitates the construction of a unified knowledge base across silos. In contrast, FedAvg may learn inconsistent inter-class relationships between different sources (e.g., the orange and pink classes in clients 1 and 2). In client 3, the coverage of different class representations is high, and the classification boundaries are not clearly defined. These observations demonstrate the effectiveness of the DEB module.

\section{Conclusions and Future Work}
In this paper, we address the problem of overfitting to specific backgrounds and the ill-posed aggregation issues caused by attribute skew in federated learning. We propose \methodname{}, which comprehensively analyzes the factors interfering FL model inference. The key idea is to generate counterfactual samples to mitigate the interference of image backgrounds. Moreover, it promotes the consistency of outputs across models from different FL client through debiasing learning. 
Experimental results demonstrate that \methodname{} significantly improves the performance of the global FL model in OOD cases.

In subsequent research, we plan to extend \methodname{} to cases where attributes cannot be directly disentangled, and develop adaptive methods for causal relationship discovery.

\section*{Acknowledgments}
This work is supported in part by the Joint Funds of the National Natural Science Foundation of China (Grant NO. U2336211); the Ministry of Education, Singapore, under its Academic Research Fund Tier 1; the RIE2025 Industry Alignment Fund – Industry Collaboration Projects (IAF-ICP) (Award I2301E0026), administered by A*STAR, as well as supported by Alibaba Group and NTU Singapore through Alibaba-NTU Global e-Sustainability CorpLab (ANGEL).

\bibliographystyle{named}
\bibliography{ijcai25}

\begin{thebibliography}{}

\bibitem[\protect\citeauthoryear{Cai \bgroup \em et al.\egroup }{2024a}]{cai2024lgfgad}
Jinyu Cai, Yunhe Zhang, and et~al.
\newblock Lg-fgad: An effective federated graph anomaly detection framework.
\newblock In {\em IJCAI}, pages 3760--3769, 2024.

\bibitem[\protect\citeauthoryear{Cai \bgroup \em et al.\egroup }{2024b}]{cai2024fgad}
Jinyu Cai, Yunhe Zhang, Zhoumin Lu, and et~al.
\newblock Towards effective federated graph anomaly detection via self-boosted knowledge distillation.
\newblock In {\em MM}, pages 5537--5546, 2024.

\bibitem[\protect\citeauthoryear{Chen \bgroup \em et al.\egroup }{2023}]{chen2023federated}
Junming Chen, Meirui Jiang, and et~al.
\newblock Federated domain generalization for image recognition via cross-client style transfer.
\newblock In {\em WACV}, pages 361--370, 2023.

\bibitem[\protect\citeauthoryear{Chen \bgroup \em et al.\egroup }{2024}]{Fedheal}
Yuhang Chen, Wenke Huang, and Mang Ye.
\newblock Fair federated learning under domain skew with local consistency and domain diversity.
\newblock In {\em CVPR}, pages 12077--12086, 2024.

\bibitem[\protect\citeauthoryear{de Luca and et al.}{2022}]{de2022mitigating}
Artur~Back de~Luca and et~al.
\newblock Mitigating data heterogeneity in federated learning with data augmentation.
\newblock {\em arXiv preprint arXiv:2206.09979}, 2022.

\bibitem[\protect\citeauthoryear{Fan \bgroup \em et al.\egroup }{2025}]{fan2025ten}
Tao Fan, Hanlin Gu, et~al.
\newblock Ten challenging problems in federated foundation models.
\newblock {\em IEEE Transactions on Knowledge and Data Engineering}, 2025.

\bibitem[\protect\citeauthoryear{Fu \bgroup \em et al.\egroup }{2025a}]{fu2025Beyond}
Lele Fu, Sheng Huang, and et~al.
\newblock Beyond federated prototype learning: Learnable semantic anchors with hyperspherical contrast for domain-skewed data.
\newblock In {\em AAAI}, pages 16648--16656, 2025.

\bibitem[\protect\citeauthoryear{Fu \bgroup \em et al.\egroup }{2025b}]{fu2025Alignments}
Lele Fu, Sheng Huang, Yanyi Lai, Chuanfu Zhang, and et~al.
\newblock Federated domain-independent prototype learning with alignments of representation and parameter spaces for feature shift.
\newblock {\em IEEE Transactions on Mobile Computing}, pages 1--16, 2025.

\bibitem[\protect\citeauthoryear{Guo \bgroup \em et al.\egroup }{2023}]{fediir}
Yaming Guo, Kai Guo, Xiaofeng Cao, and et~al.
\newblock Out-of-distribution generalization of federated learning via implicit invariant relationships.
\newblock In {\em ICML}, pages 11905--11933. PMLR, 2023.

\bibitem[\protect\citeauthoryear{He \bgroup \em et al.\egroup }{2016}]{he2016deep}
Kaiming He, Xiangyu Zhang, Shaoqing Ren, and Jian Sun.
\newblock Deep residual learning for image recognition.
\newblock In {\em CVPR}, pages 770--778, 2016.

\bibitem[\protect\citeauthoryear{Hu \bgroup \em et al.\egroup }{2023}]{hu2023gitfl}
Ming Hu, Zeke Xia, Dengke Yan, and et~al.
\newblock Gitfl: Uncertainty-aware real-time asynchronous federated learning using version control.
\newblock In {\em RTSS}, pages 145--157. IEEE, 2023.

\bibitem[\protect\citeauthoryear{Hu \bgroup \em et al.\egroup }{2024}]{hu2024fedmut}
Ming Hu, Zeke Xia, Dengke Yan, and et~al.
\newblock Fedmut: Generalized federated learning via stochastic mutation.
\newblock In {\em AAAI}, volume~38, pages 12528--12537, 2024.

\bibitem[\protect\citeauthoryear{Huang \bgroup \em et al.\egroup }{2023}]{fpl}
Wenke Huang, Mang Ye, and et~al.
\newblock Rethinking federated learning with domain shift: A prototype view.
\newblock In {\em CVPR}, pages 16312--16322. IEEE, 2023.

\bibitem[\protect\citeauthoryear{Huang \bgroup \em et al.\egroup }{2025}]{huang2025Coordinator}
Sheng Huang, Lele Fu, Yuecheng Li, Chuan Chen, and et~al.
\newblock A cross-client coordinator in federated learning framework for conquering heterogeneity.
\newblock {\em IEEE Transactions on Neural Networks and Learning Systems}, 36(5):8828--8842, 2025.

\bibitem[\protect\citeauthoryear{Li \bgroup \em et al.\egroup }{2020}]{fedprox}
Tian Li, Anit~Kumar Sahu, and et~al.
\newblock Federated optimization in heterogeneous networks.
\newblock {\em Proceedings of Machine learning and systems}, 2:429--450, 2020.

\bibitem[\protect\citeauthoryear{Li \bgroup \em et al.\egroup }{2021}]{li2021model}
Qinbin Li, Bingsheng He, and Dawn Song.
\newblock Model-contrastive federated learning.
\newblock In {\em CVPR}, pages 10713--10722, 2021.

\bibitem[\protect\citeauthoryear{Liao \bgroup \em et al.\egroup }{2024}]{liao2024swiss}
Tianchi Liao, Lele Fu, Jialong Chen, and et~al.
\newblock A swiss army knife for heterogeneous federated learning: Flexible coupling via trace norm.
\newblock {\em NeurIPS}, 37:139886--139911, 2024.

\bibitem[\protect\citeauthoryear{Lin \bgroup \em et al.\egroup }{2020}]{lin2020multi}
Chuang Lin, Sicheng Zhao, Lei Meng, and Tat-Seng Chua.
\newblock Multi-source domain adaptation for visual sentiment classification.
\newblock In {\em AAAI}, volume~34, pages 2661--2668, 2020.

\bibitem[\protect\citeauthoryear{Liu and et al.}{2025}]{liu2025grounding}
Shilong Liu and et~al.
\newblock Grounding dino: Marrying dino with grounded pre-training for open-set object detection.
\newblock In {\em ECCV}, pages 38--55. Springer, 2025.

\bibitem[\protect\citeauthoryear{Liu \bgroup \em et al.\egroup }{2021}]{liu2021feddg}
Quande Liu, Cheng Chen, Jing Qin, and et~al.
\newblock Feddg: Federated domain generalization on medical image segmentation via episodic learning in continuous frequency space.
\newblock In {\em CVPR}, pages 1013--1023, 2021.

\bibitem[\protect\citeauthoryear{Luo \bgroup \em et al.\egroup }{2022}]{luo2022disentangled}
Zhengquan Luo, Yunlong Wang, Zilei Wang, and et~al.
\newblock Disentangled federated learning for tackling attributes skew via invariant aggregation and diversity transferring.
\newblock {\em arXiv preprint arXiv:2206.06818}, 2022.

\bibitem[\protect\citeauthoryear{McMahan \bgroup \em et al.\egroup }{2017}]{fedavg}
Brendan McMahan, Eider Moore, Daniel Ramage, and et~al.
\newblock Communication-efficient learning of deep networks from decentralized data.
\newblock In {\em AISTATS}, pages 1273--1282. PMLR, 2017.

\bibitem[\protect\citeauthoryear{Meng \bgroup \em et al.\egroup }{2019}]{meng2019learning}
Lei Meng, Long Chen, and et~al.
\newblock Learning using privileged information for food recognition.
\newblock In {\em MM}, pages 557--565, 2019.

\bibitem[\protect\citeauthoryear{Meng \bgroup \em et al.\egroup }{2024}]{meng2024improving}
Lei Meng, Zhuang Qi, Lei Wu, Du~Xiaoyu, and et~al.
\newblock Improving global generalization and local personalization for federated learning.
\newblock {\em IEEE Transactions on Neural Networks and Learning Systems}, 36, 2024.

\bibitem[\protect\citeauthoryear{Morafah and et al.}{2024}]{morafah2024stable}
Mahdi Morafah and et~al.
\newblock Stable diffusion-based data augmentation for federated learning with non-iid data.
\newblock {\em arXiv preprint arXiv:2405.07925}, 2024.

\bibitem[\protect\citeauthoryear{Mu \bgroup \em et al.\egroup }{2023}]{mu2023fedproc}
Xutong Mu, Yulong Shen, Ke~Cheng, Xueli Geng, and et~al.
\newblock Fedproc: Prototypical contrastive federated learning on non-iid data.
\newblock {\em Future Generation Computer Systems}, 143:93--104, 2023.

\bibitem[\protect\citeauthoryear{Nguyen \bgroup \em et al.\egroup }{2024}]{nguyen2024fisc}
Dung~Thuy Nguyen, Taylor~T Johnson, and Kevin Leach.
\newblock Fisc: Federated domain generalization via interpolative style transfer and contrastive learning.
\newblock {\em arXiv preprint arXiv:2410.22622}, 2024.

\bibitem[\protect\citeauthoryear{Park \bgroup \em et al.\egroup }{2024}]{park2024stablefdg}
Jungwuk Park, Dong-Jun Han, and et~al.
\newblock Stablefdg: style and attention based learning for federated domain generalization.
\newblock {\em NeurIPS}, 36, 2024.

\bibitem[\protect\citeauthoryear{Qi and et al.}{2024}]{qi2024attentive}
Zhuang Qi and et~al.
\newblock Attentive modeling and distillation for out-of-distribution generalization of federated learning.
\newblock In {\em ICME}, pages 1--6. IEEE, 2024.

\bibitem[\protect\citeauthoryear{Qi \bgroup \em et al.\egroup }{2022}]{qi2022clustering}
Zhuang Qi, Yuqing Wang, Zitan Chen, Ran Wang, Xiangxu Meng, and Lei Meng.
\newblock Clustering-based curriculum construction for sample-balanced federated learning.
\newblock In {\em CICAI}, pages 155--166. Springer, 2022.

\bibitem[\protect\citeauthoryear{Qi \bgroup \em et al.\egroup }{2023}]{qi2023cross}
Zhuang Qi, Lei Meng, Zitan Chen, and et~al.
\newblock Cross-silo prototypical calibration for federated learning with non-iid data.
\newblock In {\em MM}, pages 3099--3107, 2023.

\bibitem[\protect\citeauthoryear{Qi \bgroup \em et al.\egroup }{2024}]{qi2024crosstraining}
Zhuang Qi, Lei Meng, Weihao He, Ruohan Zhang, and et~al.
\newblock Cross-training with multi-view knowledge fusion for heterogenous federated learning.
\newblock {\em arXiv preprint arXiv:2405.20046}, pages 1--12, 2024.

\bibitem[\protect\citeauthoryear{Qi \bgroup \em et al.\egroup }{2025a}]{qi2025cross}
Zhuang Qi, Lei Meng, and et~al.
\newblock Cross-silo feature space alignment for federated learning on clients with imbalanced data.
\newblock In {\em AAAI}, pages 19986--19994, 2025.

\bibitem[\protect\citeauthoryear{Qi \bgroup \em et al.\egroup }{2025b}]{qi2025global}
Zhuang Qi, Runhui Zhang, and et~al.
\newblock Global intervention and distillation for federated out-of-distribution generalization.
\newblock In {\em ICME}, pages 1--6, 2025.

\bibitem[\protect\citeauthoryear{Ren \bgroup \em et al.\egroup }{2025}]{ren2025advances}
Chao Ren, Han Yu, Hongyi Peng, Xiaoli Tang, Bo~Zhao, Liping Yi, et~al.
\newblock Advances and open challenges in federated foundation models.
\newblock {\em IEEE Communications Surveys and Tutorials}, 2025.

\bibitem[\protect\citeauthoryear{Selvaraju \bgroup \em et al.\egroup }{2017}]{selvaraju2017grad}
Ramprasaath~R Selvaraju, Michael Cogswell, Abhishek Das, and et~al.
\newblock Grad-cam: Visual explanations from deep networks via gradient-based localization.
\newblock In {\em ICCV}, pages 618--626, 2017.

\bibitem[\protect\citeauthoryear{Shenaj \bgroup \em et al.\egroup }{2023}]{shenaj2023learning}
Donald Shenaj, Eros Fan{\`\i}, Marco Toldo, and et~al.
\newblock Learning across domains and devices: Style-driven source-free domain adaptation in clustered federated learning.
\newblock In {\em WACV}, pages 444--454, 2023.

\bibitem[\protect\citeauthoryear{Shi \bgroup \em et al.\egroup }{2024}]{10336535}
Yujun Shi, Jian Liang, Wenqing Zhang, Chuhui Xue, and et~al.
\newblock Understanding and mitigating dimensional collapse in federated learning.
\newblock {\em IEEE Transactions on Pattern Analysis and Machine Intelligence}, 46(5):2936--2949, 2024.

\bibitem[\protect\citeauthoryear{{\'S}lazyk \bgroup \em et al.\egroup }{2022}]{slazyk2022cxr}
Filip {\'S}lazyk, Przemys{\l}aw Jab{\l}ecki, Aneta Lisowska, Maciej Malawski, and et~al.
\newblock Cxr-fl: deep learning-based chest x-ray image analysis using federated learning.
\newblock In {\em ICCS}, pages 433--440. Springer, 2022.

\bibitem[\protect\citeauthoryear{Sun \bgroup \em et al.\egroup }{2023}]{sun2023feature}
Yuwei Sun, Ng~Chong, and Hideya Ochiai.
\newblock Feature distribution matching for federated domain generalization.
\newblock In {\em ACML}, pages 942--957. PMLR, 2023.

\bibitem[\protect\citeauthoryear{Van~der Maaten and Hinton}{2008}]{van2008visualizing}
Laurens Van~der Maaten and Geoffrey Hinton.
\newblock Visualizing data using t-sne.
\newblock {\em J MACH LEARN RES}, 9(11), 2008.

\bibitem[\protect\citeauthoryear{Wang and Tang}{2024}]{wang2024fedccrl}
Xinpeng Wang and Xiaoying Tang.
\newblock Fedccrl: Federated domain generalization with cross-client representation learning.
\newblock {\em arXiv preprint arXiv:2410.11267}, 2024.

\bibitem[\protect\citeauthoryear{Wang \bgroup \em et al.\egroup }{2021}]{NICO_Causal}
Tan Wang, Chang Zhou, Qianru Sun, and Hanwang Zhang.
\newblock Causal attention for unbiased visual recognition.
\newblock In {\em CVPR}, pages 3091--3100, 2021.

\bibitem[\protect\citeauthoryear{Wang \bgroup \em et al.\egroup }{2022}]{wang2022meta}
Yuqing Wang, Xiangxian Li, and et~al.
\newblock Meta-causal feature learning for out-of-distribution generalization.
\newblock In {\em ECCV}, pages 530--545. Springer, 2022.

\bibitem[\protect\citeauthoryear{Wang \bgroup \em et al.\egroup }{2023}]{wang2023dafkd}
Haozhao Wang, Yichen Li, Wenchao Xu, and et~al.
\newblock Dafkd: Domain-aware federated knowledge distillation.
\newblock In {\em CVPR}, pages 20412--20421, 2023.

\bibitem[\protect\citeauthoryear{Wang \bgroup \em et al.\egroup }{2024}]{wanghaozhao1}
Haozhao Wang, Haoran Xu, Yichen Li, and et~al.
\newblock Fedcda: Federated learning with cross-rounds divergence-aware aggregation.
\newblock In {\em ICLR}, 2024.

\bibitem[\protect\citeauthoryear{Wei and Han}{2024}]{MCGDM}
Yikang Wei and Yahong Han.
\newblock Multi-source collaborative gradient discrepancy minimization for federated domain generalization.
\newblock In {\em AAAI}, volume~38, pages 15805--15813, 2024.

\bibitem[\protect\citeauthoryear{Xu \bgroup \em et al.\egroup }{2023}]{xu2023federated}
Qinwei Xu, Ruipeng Zhang, Ya~Zhang, Yi-Yan Wu, and et~al.
\newblock Federated adversarial domain hallucination for privacy-preserving domain generalization.
\newblock {\em IEEE Transactions on Multimedia}, 26:1--14, 2023.

\bibitem[\protect\citeauthoryear{Yang \bgroup \em et al.\egroup }{2020}]{yang2020federated}
Qiang Yang, Lixin Fan, and Han Yu.
\newblock Federated learning: Privacy and incentive, 2020.

\bibitem[\protect\citeauthoryear{Yi \bgroup \em et al.\egroup }{2023}]{liping4}
Liping Yi, Gang Wang, and et~al.
\newblock Fedgh: Heterogeneous federated learning with generalized global header.
\newblock In {\em MM}, pages 8686--8696. {ACM}, 2023.

\bibitem[\protect\citeauthoryear{Yi \bgroup \em et al.\egroup }{2024}]{liping2}
Liping Yi, Han Yu, Chao Ren, et~al.
\newblock Federated model heterogeneous matryoshka representation learning.
\newblock In {\em NeurIPS}, 2024.

\bibitem[\protect\citeauthoryear{Yu and et al.}{2011}]{yu2011dynamic}
Han Yu and et~al.
\newblock Dynamic witness selection for trustworthy distributed cooperative sensing in cognitive radio networks.
\newblock In {\em ICCT}, pages 1--6, 2011.

\bibitem[\protect\citeauthoryear{Yu \bgroup \em et al.\egroup }{2023}]{yu2023contrastive}
Xinhui Yu, Dan Wang, Martin~J. McKeown, and et~al.
\newblock Contrastive-enhanced domain generalization with federated learning.
\newblock {\em IEEE Transactions on Artificial Intelligence}, 5(4):1525--1532, 2023.

\bibitem[\protect\citeauthoryear{Zhang and et al.}{2024}]{zhang2024enabling}
Jiayuan Zhang and et~al.
\newblock Enabling collaborative test-time adaptation in dynamic environment via federated learning.
\newblock In {\em KDD}, pages 4191--4202, 2024.

\bibitem[\protect\citeauthoryear{Zhang \bgroup \em et al.\egroup }{2024}]{zhang2024federated}
Xunzheng Zhang, Juan~Marcelo Parra-Ullauri, Shadi Moazzeni, and et~al.
\newblock Federated analytics with data augmentation in domain generalization towards future networks.
\newblock {\em IEEE Transactions on Machine Learning in Communications and Networking}, 2024.

\bibitem[\protect\citeauthoryear{Zhang \bgroup \em et al.\egroup }{2025a}]{zhangopenprm}
Kaiyan Zhang, Jiayuan Zhang, Haoxin Li, and et~al.
\newblock Openprm: Building open-domain process-based reward models with preference trees.
\newblock In {\em ICLR}, 2025.

\bibitem[\protect\citeauthoryear{Zhang \bgroup \em et al.\egroup }{2025b}]{zhang2025federated}
Runhui Zhang, Sijin Zhou, and Zhuang Qi.
\newblock Federated out-of-distribution generalization: A causal augmentation view.
\newblock {\em arXiv preprint arXiv:2504.19882}, 2025.

\bibitem[\protect\citeauthoryear{Zhao \bgroup \em et al.\egroup }{2023}]{zhao2023federated}
Zhuang Zhao, Feng Yang, and et~al.
\newblock Federated learning based on diffusion model to cope with non-iid data.
\newblock In {\em PRCV}, pages 220--231. Springer, 2023.

\bibitem[\protect\citeauthoryear{Zhou \bgroup \em et al.\egroup }{2023}]{zhou2023fedfa}
Tailin Zhou, Jun Zhang, and Danny~HK Tsang.
\newblock Fedfa: Federated learning with feature anchors to align features and classifiers for heterogeneous data.
\newblock {\em IEEE Transactions on Mobile Computing}, 2023.

\bibitem[\protect\citeauthoryear{Zhu \bgroup \em et al.\egroup }{2024}]{zhu2024dualfed}
Guogang Zhu, Xuefeng Liu, Jianwei Niu, and et~al.
\newblock Dualfed: enjoying both generalization and personalization in federated learning via hierachical representations.
\newblock In {\em MM}, pages 11060--11069, 2024.

\end{thebibliography}

\end{document}